\pdfoutput=1

\documentclass[11pt]{article}
\usepackage[table]{xcolor}
\definecolor{lightgray}{gray}{0.8}
\usepackage[final]{acl}

\usepackage{times}
\usepackage{latexsym}

\usepackage[T1]{fontenc}

\usepackage[utf8]{inputenc}

\usepackage{microtype}

\usepackage{inconsolata}

\usepackage{graphicx}
\usepackage{tcolorbox}
\usepackage{multirow}
\usepackage{multicol}
\usepackage{booktabs}

\title{Have LLMs Reopened the Pandora's Box of AI-Generated Fake News?} 

\author{
  \textbf{Xinyu Wang\textsuperscript{1}},
  \textbf{Wenbo Zhang\textsuperscript{1}},
  \textbf{Sai Koneru\textsuperscript{1}},
  \textbf{Hangzhi Guo\textsuperscript{1}},\\
  \textbf{Bonam Mingole\textsuperscript{1}},
  \textbf{S. Shyam Sundar\textsuperscript{2, 3}},
  \textbf{Sarah Rajtmajer\textsuperscript{1}},
  \textbf{Amulya Yadav\textsuperscript{1}}
\\
  \textsuperscript{1}College of Information Sciences and Technology, Pennsylvania State University, USA
  \\
  \textsuperscript{2}Bellisario College of Communications, Pennsylvania State University, USA
  \\
  \textsuperscript{3}Department of Immersive Media Engineering, Sungkyunkwan University, South Korea
\\
\texttt{\{xzw5184, wjz5120, sdk96, hangz\}@psu.edu}
\\
\texttt{\{bjm6940, sss12, smr48, amulya\}@psu.edu}
}

\usepackage{enumitem,amssymb}
\usepackage{float}
\usepackage{booktabs}
\usepackage{makecell}
\newlist{todolist}{itemize}{2}
\setlist[todolist]{label=$\square$}
\usepackage{pifont}

\begin{document}
\maketitle
\begin{abstract}
With the rise of AI-generated content spewed at scale from large language models (LLMs), genuine concerns about the spread of fake news have intensified. The perceived ability of LLMs to produce convincing fake news at scale poses new challenges for both human and automated fake news detection systems. To address this gap, this paper presents the findings from a university-level competition that aimed to explore how LLMs can be used by humans to create fake news, and to assess the ability of human annotators and AI models to detect it. A total of 110 participants used LLMs to create 252 unique fake news stories, and 84 annotators participated in the detection tasks. Our findings indicate that LLMs are $\sim$68\% more effective at detecting real news than humans. However, for fake news detection, the performance of LLMs and humans remains comparable ($\sim$60\% accuracy). Additionally, we examine the impact of visual elements (e.g., pictures) in news on the accuracy of detecting fake news stories. Finally, we also examine various strategies used by fake news creators to enhance the credibility of their AI-generated content. This work highlights the increasing complexity of detecting AI-generated fake news, particularly in collaborative human-AI settings. 

\end{abstract}

\section{Introduction}
The vast amount of information available on the web has made it increasingly difficult to assess the credibility of online content, especially online news \cite{chung2012exploring, keshavarz2014credible}. This challenge is further complicated by the presence of highly motivated actors who create and spread fake news for various purposes, including political propaganda \cite{sanovich2017computational}. In recent years, fake news creators have increasingly turned to AI tools for creating deceptive and persuasive fake news content at scale~\cite{shu2017fake}. Within this context, the rapid rise of Large Language Models (LLMs) represents the latest (and perhaps most dangerous) addition to the fake news creator toolkit. For example, it is argued that given their advanced reasoning abilities, LLMs can easily be leveraged by motivated fake news creators to generate deceptive news content at scale \cite{allen2020evaluating}.

It is therefore important to investigate whether LLMs indeed represent an algorithmic version of Pandora’s box—tools that enable fake news creators to generate vast amounts of highly persuasive fake news content at lightning speed, which would lead to significant negative consequences for modern-day societies. At the same time, it is also important to understand whether state-of-the-art LLMs can be used to close this algorithmic Pandora’s box, i.e., whether LLMs can also serve as a solution, helping to mitigate the problems they create. For example, LLMs could play a critical role in powering the next generation of fake news detection systems, offering new ways to identify and counter deceptive content. Finally, in an effort to provide insights to future developers of next-gen fake news detection systems, it is very important to understand the strategies that can be effectively used by motivated fake news creators to generate fake news using LLMs. Concretely, we aim to answer the following research questions in this paper:

\noindent \textbf{Q1}: How easy (or difficult) is it for humans to identify LLM-generated fake news stories (as compared to real stories)? Is this task of fake/real news detection easier for state-of-the-art LLMs as compared to humans?

\noindent \textbf{Q2}: What impact do visual elements have on the detection of LLM-generated fake news?

\noindent \textbf{Q3}: From the perspective of fake news creators, which news topics and strategies are most effective in increasing the plausibility of text-based LLM-generated fake news? 

In this paper, we answer these research questions by presenting the findings from a university-level competition organized by Penn State's \href{https://csrai.psu.edu}{Center for Socially Responsible AI}, which challenged faculty, students, and staff affiliated with the university to use LLMs to generate fake news stories (\emph{creation phase}), which would then be annotated as fake/real by a different set of participants from the university (\emph{detection phase}). Our results indicate that LLMs (as detectors) are 68\% more effective than humans at identifying real news, whereas humans and LLMs perform similarly in detecting fake news ($\sim$60\% accuracy), which suggests that LLMs are not highly effective at closing the algorithmic Pandora's box of fake news.
Furthermore, we discover that LLMs find it challenging to detect fake news corresponding to certain news topics (e.g., local context in news), suggesting opportunities for targeted improvements and strategic use of models based on their strengths in specific areas. Additionally, the use of mixed prompting strategies by creators for fake news generation complicates detection for both humans and LLMs, underscoring the need for more sophisticated detection methods.

\section{Related Work}

\noindent \textbf{Detecting human-generated fake news.}
\noindent Many definitions of fake news have been proposed in the literature. Here, we adopt the definition of Allcot and Gentzkow, which refers to news articles that are intentionally and verifiably false and meant to mislead readers \cite{allcott2017social}.

Prior work on fake news detection can be organized into three approaches.
\textit{(i) Linguistic-based} to identify lexical, grammatical, and psychological features of fake news \cite{conroy2015automatic, zhang2018structured, mahyoob2020linguistic, aich2022demystifying}. \textit{(ii) Network-based} for tracking social engagements and modeling the social context of fake news \cite{conroy2015automatic, shu2017fake, wu2023decor}.
\textit{(iii) Knowledge-based} which involves manual/automatic fact-checking and source credibility validation \cite{zhou2020survey}.

\noindent \textbf{Detecting LLM-generated fake news.}
The rise of LLMs has made fake news generation alarmingly easy \cite{kreps2022all, xu2023combating, pan2023risk}, with LLMs outperforming humans in crafting convincing false narratives \cite{zhao2023more}. Whether through basic prompts, chain-of-thought techniques, or adversarial attacks \cite{wang2023implementing, zou2023universal}, LLM-generated misinformation proves to be more challenging for humans to detect than that created by humans \cite{chen2023can}. At the same time, LLMs also represent an opportunity to advance fake news detection \cite{chen2023combating, lucas2023fighting}. However, current state-of-the art fake news detection models perform poorly when faced with LLM-generated fake news \cite{wu2024fake}, and even LLMs themselves may not be able to reliably identify their own fake news \cite{jiang2024disinformation, jiang2024catching}, signaling that they are not yet ready to play a meaningful role in the evolving battle against AI-generated fake news \cite{da2024llm}.

\noindent \textbf{Human-LLM Collaboration for Fake News Generation.}
More recent techniques involve a collaboration between humans and LLMs to generate more coherent fake news stories, e.g., \cite{su2023adapting} create fake news that include both human-written and machine-generated real and fake content. \cite{sun2023med} use LLMs to add fake sentences to real articles. \cite{jiang2024disinformation} prompt ChatGPT to merge one real and one fake human-written article. \cite{pan2023risk} modify real articles by using LLMs to insert incorrect answers to related questions. However, these studies do not evaluate the ability of humans to detect fake news generated by LLMs in collaboration with humans.
Hence, in this paper, we compare the ability of both LLMs and humans to detect fake news generated through human-LLMs collaboration.

\section{Competition Design \& Details}
We hosted a four-week university-wide \href{https://csrai.psu.edu/initiatives/fake-a-thon}{competition} in Spring 2024, open to students, staff, and faculty of Penn State University. This competition aimed to engage participants in critically examining LLM-powered fake news generation and detection. Participants, recruited through targeted outreach, joined voluntarily due to their strong interest in GenAI and its impact. The competition followed a two-phase experimental design, approved by the university's Institutional Review Board (IRB):

\noindent \textbf{Phase 1: Fake News Generation.} Participants were invited to use LLMs, e.g., ChatGPT, Microsoft Copilot, to create and submit fake news stories, which could also optionally include visual elements (e.g., pictures). 
Further, participants were also required to submit a document which: (1) described their process for fake story creation, including the specific LLM used, and (2) explained how their story qualifies as fake news. This approach ensured that all submitted stories were verifiably fake. Phase 1 led to the collection of 252 fake news entries, out of which 63 contained visual elements. An expert panel selected three winners based on the persuasiveness and impact of their stories, awarding them \$500, \$300, and \$200 in prize money.

\noindent \textbf{Phase 2: Fake News Detection.} A new group of participants (not involved in Phase 1) were recruited, each of whom was asked to analyze a curated set of 18 stories: 9 fake news entries from Phase 1 and 9 real articles from a corpus of 35 real stories (details of the corpus can be found in Section~\ref{sec:real_news} of the Appendix). The 9 fake stories were randomly selected (out of 252), with each story assigned to three participants for distinct annotations. Without relying on the Internet, participants were tasked with identifying whether each of the stories was fake or real. Phase 2 involved 84 participants, each of whom annotated 18 stories. The top four annotators, who correctly identified the most stories, were awarded \$50 each. The demographic information of the annotators and recruitment messages are presented in Table~\ref{tab:statistics} and Section~\ref{sec:recruitment} in the Appendix, respectively.

\section{Annotator Performance in Phase 2}
\label{sec:annotation}
We analyze whether LLMs truly represent an algorithmic Pandora’s box - we do that by assessing Phase 2 participants' ability to accurately distinguish between real and LLM-generated fake news (from Phase 1). Furthermore, we assess whether LLMs' detection capabilities exceed human levels (if so, LLM-based detectors can enable countermeasures against LLM-generated fake news).
We present a comprehensive evaluation of LLMs on the Phase 2 detection task by benchmarking their performance against human annotators. Our analysis employs GPT-4o, a state-of-the-art multi-modal model \footnote{https://openai.com/index/hello-gpt-4o/} as an LLM annotator for the detection task in Phase 2.
To ensure a fair comparison between human and LLM annotators, we implemented two distinct processing methodologies for GPT-4o (see Tables  \ref{tab:batch_prompt} \& \ref{tab:single_prompt} in the Appendix for example prompts):

\noindent \textit{Single Processing (GPT-4o Single)}: Stories were processed individually by GPT-4o and aligned with human-annotated copies. 
For each story, GPT-4o generated annotations at three temperature settings (0.1, 0.3, 0.5) to introduce variability. A majority vote determined the final label, enhancing the reliability and robustness of the annotations.

\noindent \textit{Batch Processing (GPT-4o Batch)}: Stories were presented to GPT-4o in the same sequence as human annotators, thereby replicating the contextual flow and narrative continuity. This setting for GPT-4o mirrors the conditions under which human annotators operate. We created 84 unique batches, each matching the batches given to human participants, with 18 articles per batch input into GPT-4o. 
For the articles with visual elements, the corresponding visual element was provided at the end of the article. Temperature settings remained consistent with those used in single processing. 

\begin{figure}[h!]
    \centering
    \includegraphics[width=\linewidth]{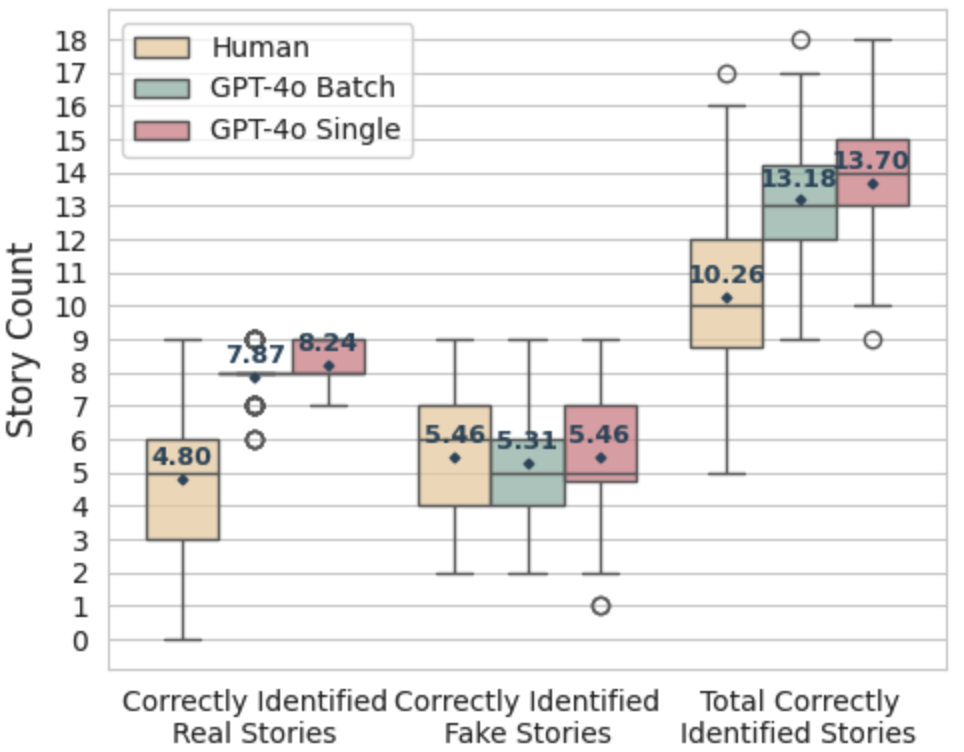}
    \caption{Box-plot comparison of correctly identified real, fake, and total stories by humans and GPT-4o.}
    \label{fig:comparison_rf}
\end{figure}

\begin{figure}[H]
    \centering
    \includegraphics[width=\linewidth]{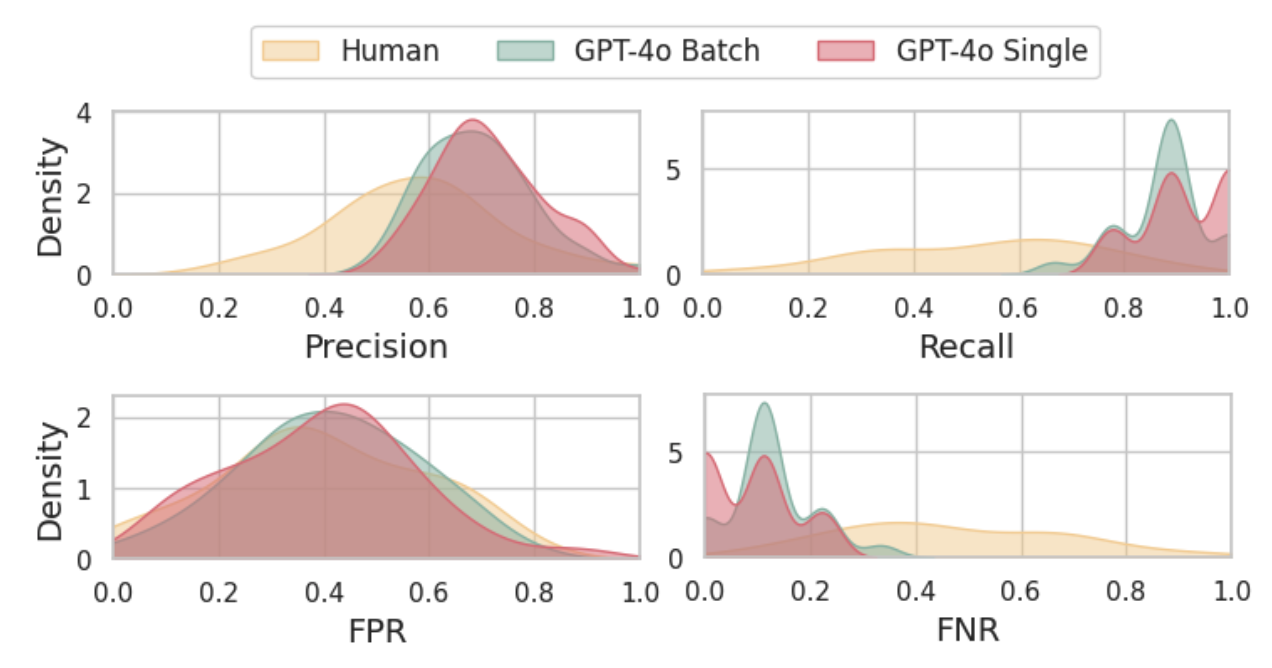}
    \caption{Density plots comparing the performance of humans and GPT-4o models (batch and single modes) across four metrics.  Top-left: Precision; Top-right: Recall; Bottom-left: False Positive Rate (FPR); Bottom-right: False Negative Rate (FNR).  }
    \label{fig:metrics}
\end{figure}

\noindent \textbf{Human VS. GPT-4o Performance.} Figure~\ref{fig:comparison_rf} shows box-plots and mean values of correctly identified real stories, correctly identified fake stories, and total correctly identified stories by human annotators, GPT-4o Batch, and GPT-4o Single.
Overall, human annotators perform worse than GPT-4o, regardless of whether GPT-4o utilizes batch or single processing methods ($p-value=9.14e^{-14}
$ and $7.54e^{-18}$, Row 2 and 4 in Table~\ref{tab:pairwise}, discussed in detail later). Importantly, this performance difference is not attributable to an inability to correctly detect fake stories, as there is no significant disparity in fake news detection between humans and GPT-4o. Instead, the discrepancy arises entirely from humans' reduced ability to accurately identify real news (p-value=$6.69e^{-27}$ and $2.97e^{-31}$, Row 8 and 10 in Table~\ref{tab:pairwise}). 
This outcome is expected, given that GPT-4o is trained on up-to-date online data sources, thereby enhancing its capacity to effectively identify authentic news.

Within human annotations, we observe no significant difference in the ability to identify real versus fake news (p-value=0.09, Row 12 in Table~\ref{tab:pairwise}). This uniform performance may be attributed to the characteristics of the selected real news, which require the same level of critical evaluation as fake ones. However, humans showed overall tendency to annotate the stories as fake, likely reflecting increased skepticism. The forewarning provided to participants about the context of the competition (e.g., they were told that they have to identify which stories are fake/real in Phase 2, which alerted them to the potential presence of fake stories) may have contributed to this effect, as prior research suggests that forewarning can reduce reliance on misinformation \cite{altay2024media}.

\begin{table*}[h!]
\centering
\small
\begin{tabular}{ccccccc}
\toprule
 Source       & SS       & DF1 & DF2 & MS       & F        & p-value  \\ \midrule
 Source       & 288.65   & 2   & 249 & 144.33   & 63.71    & \cellcolor{lightgray}4.52e-23                \\ 
 Authenticity & 304.89   & 1   & 249 & 304.89   & 136.04   & \cellcolor{lightgray}2.28e-25             \\ 
 Interaction  & 312.08   & 2   & 249 & 156.04   & 69.63    & \cellcolor{lightgray}9.60e-25             \\ \bottomrule
\end{tabular}
\caption{Mixed ANOVA analysis results showing the effects of source (Humans, GPT-4o batch, GPT-4o single), authenticity (real vs. fake), and their interaction on the detection task. The table provides: sum of squares (SS), degrees of freedom (DF1 and DF2), mean squares (MS), F-statistic (F), and p-values.} 
\label{tab:anova}
\end{table*}

In addition, Figure \ref{fig:comparison_rf} shows that there is no significant difference in the average number of correctly identified news stories between batch vs single processing for GPT-4o  (p=0.192, Row 3 in Table~\ref{tab:pairwise}). However, GPT-4o demonstrated higher accuracy in correctly identifying real news stories during single story processing compared to batch processing (p=0.008, Row 9 in Table~\ref{tab:pairwise}). This suggests that batch processing a mix of real and fake stories may impair GPT-4o's detection abilities.

To further support these results, we visualized the distribution of key performance metrics—precision, recall, false positive rate (FPR), and false negative rate (FNR), —achieved by Human, GPT-4o Batch, and GPT-4o Single in Figure~\ref{fig:metrics} (positive: real, negative: fake). We observe that the overall performance for human annotation is relatively low (in yellow), characterized by low precision and recall. Consistent with our earlier observations on the impact of forewarning, the resulting heightened skepticism among Phase 2 participants leads them to overclassify real news stories as fake (high false negative rate). Consequently, while humans may appear to detect fake news more effectively due to comparable detection rates to LLMs, this may not necessarily reflect a superior ability to detect falsity. Instead, it underscores the effect of forewarning on annotation behaviors, resulting in the misclassification of real news as fake. 

Next, we wanted to understand whether the findings gleaned from Figures \ref{fig:comparison_rf} \& \ref{fig:metrics} are statistically significant. To this end, we analyzed the performance differences between human and GPT-4o annotators by conducting a mixed analysis of variance (ANOVA) \cite{murrar2018mixed}. Our analysis framework included:
\noindent \textit{Between-Subjects Factor}: Source of annotation, categorized into three levels—human annotators, GPT-4o Batch, and GPT-4o Single.
\noindent \textit{Within-Subjects Factor}: Authenticity of the stories, categorized as fake or real.

Table~\ref{tab:anova} summarizes the mixed ANOVA results, which reveal a significant overall difference among the three annotation sources and between the authenticity of the news (p-value =$4.52e^{-23}$ and $2.28e^{-25}$ ). Additionally, there is a significant interaction effect (p-value=$9.6e^{-25}$), indicating that the impact of authenticity (correct identification of real vs. fake news) varies across the different annotation sources (human processing vs. GPT-4o batch processing vs. GPT-4o single processing).

Given the significant interaction effects, we conducted simple main effects analyses for pairwise comparisons to further examine these differences. Detailed statistics for pairwise comparisons are provided in Table~\ref{tab:pairwise} in the Appendix, where we also re-affirm our findings using non-parametric tests (See Section~\ref{sec:robustness} and Table~\ref{tab:pairwise_non_parametric} in the Appendix) for all pairwise comparisons.

\begin{tcolorbox}[boxsep=1pt,left=3pt,right=3pt,top=3pt,bottom=3pt]
\textbf{Finding: Fake News Detection Remains Challenging for Both Humans and LLMs.}
Although LLMs perform 68\% better on average at identifying real news largely due to their extensive training on real content, this advantage does not carry over to fake news detection. With an average accuracy rate of only 60\% for fake news detection, LLMs demonstrate subpar performance and fall short of being reliable tools for combating misinformation. These findings highlight a significant limitation: neither LLMs nor humans alone are adequately equipped to tackle the complex challenge of fake news detection. 

\end{tcolorbox}

\begin{table*}
\centering
\small
\begin{tabular}{c|c|c|c|c|c}
\toprule
 && \textbf{\makecell{w/ VE\\ (VE+Text)}} &\textbf{\makecell{w/ VE\\(VE->Text)}}&\textbf{\makecell{w/ VE\\ (Text->VE)}}& \textbf{\makecell{w/o VE}}\\  
\midrule
\multirow{3}{*}{GPT-4o} & $VE_{AI}$& \cellcolor{green!30}71.43\% &50.00\% &\cellcolor{green!30}75.00\% & \multirow{3}{*}{59.26\%}\\\cline{2-5}
 & $VE_{Authentic}$&57.14\%  & \cellcolor{green!30}52.94\%&54.90\% & \\\cline{2-5}
  & $VE_{All}$&\textbf{65.08\%}  &52.38\% & 58.73\%& \\\midrule
\multirow{3}{*}{Gemini} &$VE_{AI}$ &\cellcolor{green!30}75.00\%&\cellcolor{green!30}66.67\%& 66.67\% &\multirow{3}{*}{68.25\%} \\\cline{2-5}
 & $VE_{Authentic}$& 60.78\% &60.78\% &\cellcolor{green!30}70.59\% & \\\cline{2-5}
  & $VE_{All}$& 63.49\% &61.90\% &\textbf{69.84\%} & \\\midrule
\multirow{3}{*}{Human Majority Voting}&$VE_{AI}$ &\multicolumn{3}{c|}{\cellcolor{green!30}75.00\%}&\multirow{3}{*}{66.67\%} \\ \cline{2-5}
 & $VE_{Authentic}$& \multicolumn{3}{c|}{64.71\%} &  \\\cline{2-5}
  & $VE_{All}$& \multicolumn{3}{c|}{\textbf{66.67\%}} & \\
\bottomrule
\end{tabular}
\caption{Performance comparison of GPT-4o, Gemini, and Human on Fake News with (w/ VE) and without Visual Elements (w/o VE). With visual elements, (VE+Text): Simultaneous Text and Image Input; (VE -> Text): Sequential Image-First Input; and (Text -> VE): Sequential Text-First Input represent three different types of information processing; w/o VE: Fake stories without any visual elements; $VE_{AI}$: Images generated by AI models;
$VE_{Authentic}$: Images sourced from the internet or other real-world sources, not AI-generated; $VE_{All}$: All images. Green indicates better performance comparing $VE_{AI}$ and $VE_{Authentic}$ for the particular processing method; Bolded indicates better performance across processing methods and the presence of VE.  }
\label{tab:VE}
\end{table*}

\section{Role of Visual elements in Detection}

In this section, we explore the influence of visual elements on the accuracy of fake news detection by humans vs LLMs. Specifically, our goal in this section is to answer the following question: Does the presence of visual elements aid in the detection of fake news stories by humans compared to LLMs? We conduct a comparative analysis between fake news instances that include visual elements and those presented with text alone. Out of the 252 fake news entries, 63 contained both text and visual elements, while 189 were text-only. 

To ensure that our comparison focuses exclusively on fake news detection and operates at the story level, we employ LLMs in single processing mode since our findings in Section \ref{sec:annotation} show that there are no significant differences in overall fake news detection performance between single vs batch processing methods. Along with GPT-4o\footnote{GPT-4o points to gpt-4o-2024-08-06}, we include Gemini\footnote{Gemini refers to Gemini-1.5-flash-001}, another multi-modal model, to further validate our results.

To simulate the diverse ways humans can process information from both text and images (e.g., whether by focusing on text first, focusing on images first, or considering both simultaneously), we conducted three sets of experiments with LLMs by varying input sequences:

\noindent \textit{Simultaneous Text \& Image Input (VE+Text)}: {Both text and image are processed together in a single prompt and provided to the model.}

\noindent \textit{Sequential Text-First Input (Text->VE)}: { 
First, the model receives the article title and content through an initial prompt (Table \ref{tab:prompt_first_text}) for an initial judgement. Next, a separate prompt (Table \ref{tab:prompt_first_text}) including image, and the chat history (prompt-1, response) was sent to the model to generate a label.}

\noindent \textit{Sequential Image-First Input (VE -> Text)}: {Here, the image along with the article title, was presented to the model through a single prompt-3 (Table \ref{tab:prompt_first_image}) to make an initial judgment. Next, prompt-4 (Table \ref{tab:prompt_first_image}) including article content and chat history was sent to the model to generate a label.}

Furthermore, the nature of images may also impact detection accuracy. Participants employed various methods to generate or source images, such as using AI tools like DALL-E or conducting image searches via Google, which likely resulted in authentic images being used for fake news. To account for this variability, we utilize a consumer-grade AI-generated image detector\footnote{\url{https://illuminarty.ai/en/}} to verify the origin of each image. In addition to the labels generated from the image detector, a co-author independently labeled each image. Any discrepancies between the detector's labels and the human labels were then reviewed and resolved by a third coder (a different co-author), ensuring accuracy in the final labels.
Cross-referencing with the generation processes provided by participants, we identified 12 AI-generated images and 51 authentic images.

Table~\ref{tab:VE} compares the accuracy across annotation sources, processing orders, and image types. The results show that images do not consistently impact detection accuracy, with both LLMs and humans showing similar performance regardless of whether visual elements are present. However, detection accuracy improves significantly for humans when the visual elements are AI-generated (75.00\% vs. 64.71\%). For LLMs, this improvement is observed consistently only when the image and text are processed simultaneously (GPT-4o: 71.43\% vs. 57.14\%; Gemini: 75\% vs. 60.78\%).

\begin{tcolorbox}[boxsep=1pt,left=3pt,right=3pt,top=3pt,bottom=3pt]
\textbf{Finding: Visual Elements Have a Modest but Inconsistent Impact on Fake News Detection.}
With the optimal processing modality, visual elements improved fake news detection accuracy by <6\% on average. 
Notably, when visual elements are AI-generated, they generally aid both humans and LLMs in detecting fake news. This suggests that AI-generated images are currently easier to identify as being AI-generated. 
\end{tcolorbox}

\section{Fake News Generation}
Finally, in an effort to provide insights to future developers of next-generation fake news detection systems, we focus on categorizing the techniques and choice of topics that were deemed to be effective at creating convincing fake news content by creators during Phase 1 of the competition.

In particular, to isolate the effects of topic selection and prompting strategies used by Phase 1 participants to create fake news, our analysis focuses exclusively on the textual components of all the 252 fake news stories that were submitted in the competition. We conduct thematic analyses that investigate both the choice of topics and the strategic techniques employed, assessing how these factors influence the effectiveness of fake news detection by human annotators and LLMs. In addition to GPT-4o and Gemini, we incorporated the open-source Llama-3.1\footnote{Llama-3.1 refers to Llama-3.1-8B-Instruct} for text-only comparisons. The models utilized by the participants in the creation process are detailed in Table~\ref{tab:tools} in the Appendix.

\subsection{Thematic Analysis}

We performed a two-stage thematic analysis (detailed in Section~\ref{sec:clustering} in the Appendix) of the fake news stories, clustering them into 8 topics:

\noindent \textbf{T1: Scientific Research (19.84\
\%)}: News related to scientific discoveries across various fields, including environmental science, earth science, astrophysics, and other areas of research.

\noindent \textbf{T2: AI and Technology (12.30\%)}: News focused on artificial intelligence and technological advancements in various sectors, including robotics, software development, and innovation in IT.

\noindent \textbf{T3: Local and Community News (12.70\%)}: News centered around events and developments in the local community or regional context where the experiment takes place.

\noindent \textbf{T4: COVID-19 and Public Health (3.57\%)}: News specifically related to the COVID-19 pandemic, public health initiatives, medical guidelines, and other health-related events.

\noindent \textbf{T5: Global Affairs (8.33\%)}: News on major international events, e.g., conflicts, diplomacy, etc. 

\noindent \textbf{T6: Politics and Policy (13.49\%)}: News focused on U.S. political developments, e.g., elections, government policies, legislation. 

\noindent \textbf{T7: Medical and Clinical Studies (7.94\%)}: News concerning research findings in the medical field, e.g., clinical trials, psychological experiments. 

\noindent \textbf{T8: Entertainment and Media (13.49\%)}: News revolving around celebrities, sports, movies, and other areas of popular culture.

The results show that most of the generated fake stories are science-related news (19.84\%). A unique aspect of this dataset is that participants frequently chose to generate local news content (12.7\%). This trend suggests a potential preference for creating fake stories that are context-specific and more relatable to the potential audience, differing from the typical broader narratives seen in conventional fake news datasets~\cite{kim2023covid}.
We subsequently calculated the detection accuracy rates for fake news within each identified topic, as presented in Table~\ref{tab:Topic}. To eliminate the influence of visual elements on detection accuracy, we based human annotation accuracy solely on the 189 text-only fake news stories. We used the same 189 text-only stories to evaluate LLM performance, ensuring directly comparable accuracy scores.

\begin{table}[!ht]
\small
\centering
\begin{tabular}{@{}c|c||c|c|c@{}}
\toprule
\textbf{Topic} & Human& GPT-4o& Gemini&Llama\\ \midrule
T1 &63.64\%    &57.57\% &\cellcolor{green!30}69.70\% &30.30\%\\ 
T2& 57.69\%     &50.00\% &\cellcolor{green!30}73.08\%&38.46\%\\ 
T3&\cellcolor{green!30}66.67\%  & 29.17\% & 45.83\% & 16.67\%\\ 
T4&71.43\%   &71.43\% &\cellcolor{green!30}85.71\%&42.86\% \\ 
T5&70.59\%     &\cellcolor{green!30}76.47\% &58.82\%&52.94\% \\ 
T6&62.96\%    &\cellcolor{green!30}66.67\% &62.96\%&48.15\% \\ 
T7&70.00\%   &70.00\% &\cellcolor{green!30}90.00\%&45.00\% \\ 
T8&72.00\%    &72.00\% &\cellcolor{green!30}76.00\%&48.00\%\\ \bottomrule
\end{tabular}
\caption{Text-based detection accuracy of fake news stories by topic. Green indicates best performance across all detectors for the particular topic.  }
\label{tab:Topic}
\end{table}

Gemini performs best on 5 out of 8 topics (T1,T2,T4,T7, and T8), whereas GPT-4o shows strong accuracy in Global Affairs (76.47\%; T5) and Politics and Policy (66.67\%; T6). However, both models fare poorly with Local News (T3), likely due to limited coverage of localized events in their training data. In contrast, Llama-3.1 consistently shows lower accuracy across all topics, while human annotators achieve moderate performance overall, with their best results in Local News (66.67\%; T3).
These differences suggest that selecting a detection model should be guided by the specific use case, as each model may perform better in certain topics. A more effective approach could involve combining multiple models to tackle topic-specific challenges in misinformation detection.

\subsection{Examination of Generation Processes}

Next, we focus on understanding the purposeful approaches individuals take when interacting with generative AI to create fake news. 
In the following sections, we thematically analyze the textual content contained within the additional document submitted by Phase 1 participants (which asks them to write a freeform description of how they used LLMs to create fake stories for the competition). The systematic thematic analysis of this textual content will enable us to identify primary prompting strategies and secondary output optimization techniques used by creators to enhance the effectiveness of misleading content, providing insights into the deliberate manipulation tactics employed.

\subsubsection{Primary Prompting Strategies}
We now discuss distinct prompting approaches used by participants to generate fake news:

\noindent \textbf{P1. Direct Instruction (26.59\%):} Participants provide a clear, concise directive to the AI, offering a simple constraint on the topic or style, such as "Can u generate a story for me that I can submit to the New York times for news".

\noindent \textbf{P2. False Statement Expansion (46.03\%):} Participants frame the story around a false statement, guiding the AI to produce content within that narrative. E.g., ``Write me a 500-word news article about how scientists have discovered how dinosaurs really sounded, include research and quotes."

\noindent \textbf{P3. Fact-driven Distortion (11.51\%):}
Participants provide a set of true facts or accurate information and prompt the AI to generate fake news that incorporates and distorts these facts, blending truth with misinformation to create a more convincing narrative. E.g, "Imagine that you are feminist journalist writing about Roe vs Wade. 
Use the information provided to write an effective article on this topic: <real statements>". 

\noindent \textbf{P4. Narrative Imitation (17.06\%):} Participants provide an existing article, or a URL, or datasets of real and fake news to instruct the AI in creating a similar but false story. This approach supplies the AI with more contextual information to enhance the accuracy of the generation.

\subsubsection{Secondary Optimization Strategies }
With the intent to deceive, participants may additionally incorporate supplementary strategies alongside the primary prompting methods to further enhance the credibility of the content.

\noindent \textbf{S1. Stylistic Adjustment (24.21\%):} Participants prompt the AI to adopt specific styles, e.g., "make it sound like a journal paper" 
to shape the style and structure of the generated content.

\noindent \textbf{S2. Authority Referencing (20.24\%):}  Participants enhance the credibility of fake news by fabricating citations and/or quotes, often attributing them to well-known figures, reputable sources (e.g., Science Journal), and even non-existent authorities.

\noindent \textbf{S3. Contextual Enhancement (7.54\%):} Participants instruct the AI models to add examples and specific details to the fake story, enhancing its contextual depth and relatability.

\noindent \textbf{S4. Post-Prompt Fact Injection (1.98\%)}: Some participants inject additional real facts into the fake news (manually or via the LLM) after the initial LLM output, making the story harder to detect.

\noindent \textbf{S5. Iterative Refinement (22.22\%):} Participants employ a sequence of prompts to iteratively refine the AI-generated content, progressively improving its quality. For instance, they might repeatedly prompt with "make it more believable."

\noindent \textbf{S6. Multiple-Output Selection (4.37\%):} Participants ask the AI to generate multiple versions of a story and then select the most convincing output.

\noindent \textbf{S7. Manual Revision (9.13\%):} Participants manually revise AI-generated content by identifying and adjusting unnatural phrases or details, e.g., fake names, unrealistic locations. This approach blends AI-generated material with human intervention, resulting in a more authentic narrative.

Figure~\ref{fig:HM} in the Appendix shows the frequency of co-usage of core prompting and secondary optimization strategies to create fake news. This figure shows that when participants intend to deceive (since they wanted to win the competition), co-usage of multiple strategies is commonly observed.

Table~\ref{tab:strategy_acc} presents the annotation accuracy of humans and LLMs based on the strategies employed during the fake news creation process. We find that GPT-4o does poorly when fake news includes rich contextual information (27.27\%; S3), such as specific examples and detailed descriptions. Gemini, on the other hand, performs relatively poorly when fake news is generated by mimicking narratives from existing news articles (57.5\%; P4). All LLMs perform poorly when post-prompt fact injection is employed (0\%; S4), even though the small sample size limits generalizability. However, manual modifications to fake news significantly deceive human judgment but are more easily identified by LLMs (GPT-4o: 76.92\%; Gemini: 84.62\%, S7) in comparison to human (38.46\%, S7). This aligns with findings in \citet{chen2023can}, who found that LLMs perform better at detecting human-written than LLM-generated misinformation.

\begin{table}[!ht]
\small
\centering
\begin{tabular}{@{}c|c||c|c|c@{}}
\toprule
\textbf{Strategy} & Human&  GPT-4o& Gemini&Llama \\ \midrule
P1&67.80\%  &62.71\% &\cellcolor{green!30}74.58\%&45.76\%\\ 
P2&65.71\%&68.57\% &\cellcolor{green!30}72.86\%&40.00\%\\ 
P3  &\cellcolor{green!30}70.59\% &47.06\%&64.71\%&29.41\%\\ 
P4 & \cellcolor{green!30}60.00\% &50.00\%&57.50\%&30.00\%  \\ 
S1&  64.58\% &58.33\% &\cellcolor{green!30}70.83\%&31.25\% \\ 
S2&63.41\% &63.41\% &\cellcolor{green!30}65.85\%&26.83\% \\ 
S3  &\cellcolor{green!30}63.64\%  &27.27\% &\cellcolor{green!30}63.64\%&36.36\% \\ 
S4 &\cellcolor{green!30}50.00\% &0.00\% &0.00\%&0.00\%  \\ 
S5&64.86\%  &56.76\% &\cellcolor{green!30}70.27\%&29.73\% \\ 
S6&66.67\%&\cellcolor{green!30}77.78\% &\cellcolor{green!30}77.78\%&22.22\%\\ 
S7 &38.46\%  &76.92\% &\cellcolor{green!30}84.62\%&23.08\%\\ \bottomrule
\end{tabular}
\caption{
Text-based detection accuracy of fake news stories by strategy. Green indicates best performance across all detectors for the particular strategy. 
} 
\label{tab:strategy_acc}
\end{table}

\begin{tcolorbox}[boxsep=1pt,left=3pt,right=3pt,top=3pt,bottom=3pt]
\textbf{Finding: Human-AI Collaboration Creates New Challenges.} Creators often use multiple strategies during prompting and output optimization in combination, and such human-AI collaboration complicates fake news detection. 
\end{tcolorbox}

\section{Conclusion and Discussion}

This study provides key insights into the performance of humans and LLMs in detecting fake news that was created with human-AI collaboration. While LLMs are generally 68\% better at identifying real news, humans who are forewarned about the presence of fake news can perform at a similar level to LLMs ($\sim$60\%). This subpar performance of LLMs at detecting fake news illustrates how the algorithmic Pandora's box of LLM driven fake news creation cannot rely on LLMs as a countermeasure for improved detection. Alternatively, these results suggest that maintaining a high level of alertness and skepticism could be beneficial when fake news is prevalent. 
However, this heightened alertness comes with potential drawbacks. Our findings suggest that excessive skepticism from forewarning can lead individuals to dismiss legitimate news. This "discounting effect" may reduce news source credibility, foster cynicism, and drive users toward personal channels like messaging apps, which are less regulated and more prone to misinformation.

Additionally, LLMs are more effective when processing news stories individually rather than in batches. This finding suggests that presenting information individually allows models to focus more precisely on each piece of content, reducing contextual interference from adjacent stories.

Visual aids in fake news detection presents another layer of complexity. With the optimal processing modality, visual elements improve fake news detection accuracy by <6\% on average. However, depending on the source and processing mode, visuals can either improve or impair detection accuracy compared to text-only content. Further research is needed to better understand the impact of visuals on fake news detection.


Finally, in text-based fake news creation, participants tend to focus on certain topics, with LLM performance varying by topic, suggesting challenges and opportunities for targeted improvements. Specific LLMs could be strategically employed for detecting fake news in areas where they perform better. Additionally, when incentivized, participants often combine strategies during both prompting and post-prompting phases, complicating detection for both humans and LLMs. As fake news creation grows more sophisticated, detection efforts must evolve to match these complex scenarios.

\section{Limitations}
The study's conclusions are based on fake news submissions provided by participants in the competition. Although the vast majority of submissions were of high quality, a very small number were of lower quality. These lower-quality submissions were retained in the dataset to replicate the variability found in real-world misinformation creation, where some creators invest substantial effort while others do not. The inclusion of these data points is expected to have minimal impact on the overall comparison, as they were equally distributed across all annotation modes. Furthermore, in the thematic analyses, these lower-quality entries were excluded from all categories to prevent any potential bias in the results.

To ensure that the news created by participants is verifiably fake, we asked them to provide reasoning for why their news is demonstrably false—an additional step to confirm its falsity. However, there remains the possibility that some news may be borderline in terms of authenticity or could potentially become true in the future.

In addition, the real news stories included in this study were intentionally selected for their unusual or distinctive characteristics to avoid incorporating well-known, mundane facts with which human annotators are likely to be already familiar. While such atypical real news stories are prevalent in actual media and may attract more attention, a substantial portion of real news remains mundane and was not included in this analysis. This selection bias may limit the generalizability of our findings, as detection performance could vary when dealing with a broader spectrum of real news types. Future research should investigate the detection accuracy of human annotators across diverse categories of real news, encompassing both unusual and mundane stories, to provide a more comprehensive assessment of human detection capabilities.

\section{Ethical Considerations}
First, there is a risk of unintentionally promoting the misuse of generative AI tools. While the study focuses on detecting AI-generated fake news, it also reveals strategies for making such content more convincing, which could be exploited by malicious actors.

Second, privacy concerns arise from the data collection through the university-wide competition. Although analyses are aggregated and IRB approval was obtained, care must be taken to protect participant identities, especially since some fake news stories contain local context. We also ensure that the generated content is restricted to research use to avoid spreading misinformation.

Third, bias in the creation and detection of fake news remains a persistent concern. AI models trained on biased data may reinforce stereotypes, and human annotators may bring their own biases, potentially affecting the fairness of judgments.

Lastly, the key ethical issue arises from human-AI collaboration in generating fake news, especially when there is an incentive to deceive. The study shows that combining AI tools with human input can produce convincing misinformation. This complicates detection efforts and presents new challenges for preserving the integrity of online information systems, which requires further attention from the academic community.

\section*{Acknowledgments}
The Fake-a-thon competition reported in this paper was sponsored by the Center for Socially Responsible Artificial Intelligence (CSRAI) at Penn State University. Co-author Sundar is supported by MSIT(Ministry of Science, ICT), Korea, under the Global Scholars Invitation Program (RS-2024-00459638). This work was also partially supported by NSF award \#2318460.

\bibliography{main.bib}

\appendix

\section{Appendix}
\label{sec:appendix}
\renewcommand{\thetable}{A\arabic{table}}
\renewcommand{\thefigure}{A\arabic{figure}}
\setcounter{table}{0} 
\setcounter{figure}{0}

\subsection{Real News Corpus Creation}
\label{sec:real_news}
The organizers of the competition carefully curated a set of 35 real news stories from reputable and well-established news outlets to ensure high standards of credibility and authenticity. These stories were sourced from widely recognized sources such as The New York Times, The Washington Post, CNN, and The New York Post. All 35 real news stories included images from the reputable sources, which were also provided as visual elements during the annotation phase.

\subsection{Recruitment Materials}
\label{sec:recruitment}
Below is the primary component of the recruitment materials provided to participants in the competition.

\textit{Purpose of the Study}: The competition is organized into two stages: the creation of fake news stories online and the detection of these fake stories among real ones in an in-person event. This is an opportunity to learn, engage, and enhance your skills in evaluating the credibility of online information.

\textit{Compensation}: For the competition, the top 3 AI-generated fake news that fools the most annotators will win \$500, 300, and 200. The top 4 annotators who identify most of the news correctly will win \$50 each. 
        
\textit{Privacy and Confidentiality}: Efforts will be made to limit the use and sharing of your personal research information to the researchers involved in this study. In the event of any publication or presentation resulting from the research, no personally identifiable information will be shared. However, absolute confidentiality cannot be guaranteed.
\begin{table}[ht]

\centering
\begin{tabular}{@{}lc@{}}
\toprule
\textbf{Characteristic}                      & \textbf{Values} \\ \midrule
\textbf{Gender}              \\
  Female                         &   20     \\
  Male                  &   63      \\
  Non-binary     &  1        \\ \addlinespace
\textbf{Occupation}               \\
Undergraduate Student       & 53         \\
Graduate Student  & 28       \\
Research Intern &   2        \\
Student Intern           &  1   \\ 
\addlinespace
\textbf{GenAI Usage}               \\
Daily&   18    \\
Very Frequently (Multiple times/week)                         &     26\\

Frequently (Once/week)                       &    18     \\
Occasionally (1-3 times/month)      &  12    \\ 
Rarely (Less than once/month)        &    9      \\ 
Never &  1       \\ 
\bottomrule
\end{tabular}

\caption{Annotator demographics (N=84).}
\label{tab:statistics}
\end{table}

\begin{table*}[]
    \centering
    \small
    \begin{tabular}{p{14.5cm}}
    \toprule
    \textbf{system\_role:} You are an expert in the fake news detection area. You are assigned to the task of fake news detection in English.\\

    \textbf{prompt\_text:} You will analyze a set of news articles and evaluate whether each one is Fake or Real. Each article is accompanied by a corresponding index under <index> tag and title under the <Title> tag and article content under the <Content> tag. After making your decision, provide an explanation that outlines the key reasons behind your conclusion. Consider any patterns or inconsistencies that stood out to you while reading.\\
    <Index> \{news\_index\}    </Index>\\
    <Title> \{article\_title\} </Title>\\
    <Content> \{article\_content\} </Content> \\
    \textit{Expected Output:}
    <Index>: [Corresponding news index]
    <Label>: [Fake/Real]
    <Explanation>: [Provide a detailed explanation based on your analysis of the article’s content, highlighting any indicators that led you to your decision.]\\
    \bottomrule
    \end{tabular}
    \caption{Prompt for batch processing.}
    \label{tab:batch_prompt}
\end{table*}

\begin{table*}[h!]
\centering
\small
\begin{tabular}{p{14.5cm}}
\toprule
\textbf{system\_role:} You are an expert in the fake news detection area. You are assigned to the task of fake news detection in English.\\
\textbf{prompt\_text:}
Task:
Read the following article title under the <Title> tag and article content under the <Content> tag.  
And determine whether the article is fake or real.
After making your decision, provide an explanation that outlines the key reasons behind your conclusion.
Consider any patterns or inconsistencies that stood out to you while reading.\\

<Title>
\{article\_title\}
</Title>

<Content> 
\{article\_content\}
</Content> 
 
\textit{Expected Output:}
<Label>: [Fake/Real]
<Explanation>: [Provide a detailed explanation based on your analysis of the article’s content, highlighting any indicators that led you to your decision.]\\
\bottomrule
\end{tabular}
\caption{Prompt for single processing.}
\label{tab:single_prompt}
\end{table*}

\begin{table*}[h!]
\centering
\small
\begin{tabular}{cccccccccc}
\toprule
Row&Contrast & Within & A         & B         &  T       & dof  &  p-unc    & p-corr    \\ \midrule
1& Authenticity  & -   & Fake & Real      & -9.38   & 251    & \cellcolor{lightgray}4.14e-18 & -  \\ 
2& Source        & -   & Batch  & Human  & 8.32    & 166  & 3.05e-14 & \cellcolor{lightgray}9.14e-14     \\ 
3& Source        & -   & Batch                & Single         & -1.87   & 166 & 0.064    & 0.192       \\ 
4& Source        & -   & Human                & Single      & -9.85   & 166   & 2.51e-18 & \cellcolor{lightgray}7.54e-18    \\ 
 \midrule
5&Authenticity * Source & Fake& Batch     & human                      & -0.59   & 166   & 0.554    & 1.0        \\ 
6&Authenticity * Source & Fake  & Batch     & Single                     & -0.63   & 166  & 0.533    & 1.0     \\ 
7&Authenticity * Source & Fake  & Human     & Single                     & 0.00    & 166   & 1.0      & 1.0 \\ 
8& Authenticity * Source & Real  & Batch     & Human        & 13.21   & 166    & 1.12e-27 & \cellcolor{lightgray}6.69e-27     \\ 
9& Authenticity * Source & Real  & Batch     & Single                    & -3.27   & 166    & 0.001   & \cellcolor{lightgray}0.008 \\ 
10& Authenticity * Source & Real  & Human     & Single                    & -14.76  & 166   & 4.95e-32 & \cellcolor{lightgray}2.97e-31  \\ \midrule
11& Source * Authenticity&Batch&	Fake&Real&-14.65 &83&9.72e-25&\cellcolor{lightgray}2.92e-24    \\ 
12& Source * Authenticity&	Human&	Fake&Real&2.21&	83& 0.030&0.090 \\ 
13& Source * Authenticity&	Single&	Fake &Real&-14.13&83 & 8.52e-24 & \cellcolor{lightgray}2.56e-23  \\ \bottomrule
\end{tabular}
\caption{Results of pairwise tests from post-hoc simple main effects analyses, comparing between different sources (human with forewarning, GPT-4 batch, GPT-4 single) and between fake and real stories. The table provides: t-statistics (T), degrees of freedom (dof), uncorrected p-values (p-unc), and corrected p-values after bonferroni correction (p-corr). Grey indicates the difference is significant. }
\label{tab:pairwise}
\end{table*}

\begin{table*}[h!]
\centering
\small
\begin{tabular}{ccccccccc}
\toprule
Contrast & Within & A         & B         &  U&W        &  p-unc    & p-corr    \\ \midrule
 Authenticity  & -   & Fake & Real  & -    & 4441.0& \cellcolor{lightgray}4.90e-17 & -  \\ 
 Source        & -   & Batch  & Human  & 5759.5    & -  & 1.03e-12& \cellcolor{lightgray}3.09e-12	     \\ 
 Source        & -   & Batch                & Single         & 2944.5   & - & 0.061   & 0.18       \\ 
 Source        & -   & Human                & Single      & 1031.5   & -  & 1.63e-15 & \cellcolor{lightgray}4.90e-15	 \\ 
 \midrule
Authenticity * Source & Fake& Batch     & human                      & 3349.0   & -   & 0.56    & 1.0        \\ 
Authenticity * Source & Fake  & Batch     & Single                     & 3306.0   & -  & 0.47    & 1.0     \\ 
Authenticity * Source & Fake  & Human     & Single                     & 3512.5    & -   & 0.96      & 1.0 \\ 
 Authenticity * Source & Real  & Batch     & Human        & 6583.0   & -    & 3.42e-23 & \cellcolor{lightgray}2.05e-22     \\ 
 Authenticity * Source & Real  & Batch     & Single                    &2616.5   & -    & 0.0097   & \cellcolor{lightgray}0.0078 \\ 
 Authenticity * Source & Real  & Human     & Single                    & 309.0  & -   & 3.70e-25 & \cellcolor{lightgray}2.22e-24  \\ \midrule
 Source * Authenticity&Batch&	Fake&Real&-	 &14.0	&4.43e-14&\cellcolor{lightgray}1.33e-13  \\ 
 Source * Authenticity&	Human&	Fake&Real&-&949.5& 0.058&0.17 \\ 
 Source * Authenticity&	Single&	Fake &Real&-&14.0 & 4.66e-14 & \cellcolor{lightgray}1.40e-13  \\ \bottomrule
\end{tabular}
\caption{Results of non-parametric pairwise tests from post-hoc simple main effects analyses, comparing between different sources (human with forewarning, GPT-4 batch, GPT-4 single) and between fake and real stories. The table provides: U-values (U), W-values (W), uncorrected p-values (p-unc), and corrected p-values after bonferroni correction (p-corr).}
\label{tab:pairwise_non_parametric}
\end{table*}

\begin{table*}[h!]
\centering
\small
\begin{tabular}{p{14.5cm}}
\toprule
\textbf{system\_role:} You are an expert in the fake news detection area. You are assigned to the task of fake news detection in English.\\
\textbf{prompt1\_text:}
Please read article title under the <Title> tag and article content under the <Content> tag. 
Do you think this article is fake or real?
    
<Title>
\{article\_title\}
</Title>

<Content> 
\{article\_content\}
</Content>

\textbf{prompt2\_text:}
Task:
There is also an image associated with this article. Understand the information from the image.
Combine the information from both text and image. 
And determine whether the article is fake or real.
After making your decision, provide an explanation that outlines the key reasons behind your conclusion.
Consider any patterns or inconsistencies that stood out to you while reading.
 
\textit{Expected Output:}
<Label>: [Fake/Real]
<Explanation>: [Provide a detailed explanation based on your analysis of the article’s content, highlighting any indicators that led you to your decision.]\\
\bottomrule
\end{tabular}
\caption{Prompt for experiment of the Sequential Text-First.}
\label{tab:prompt_first_text}
\end{table*}

\begin{table*}[h!]
\centering
\small
\begin{tabular}{p{14.5cm}}
\toprule
\textbf{system\_role:} You are an expert in the fake news detection area. You are assigned to the task of fake news detection in English.\\
\textbf{prompt3\_text:}
Based on the article title under the <Title> tag and image information, Do you think this article is fake or real?

<Title>
\{article\_title\}
</Title>

\textbf{prompt4\_text:}
Task:
Right now, continously read the article content under the <Content> tag. 
Combine the information from both text and image. 
And determine whether the article is fake or real.
After making your decision, provide an explanation that outlines the key reasons behind your conclusion.
Consider any patterns or inconsistencies that stood out to you while reading.

<Content> 
\{article\_content\}
</Content> 
 
\textit{Expected Output:}
<Label>: [Fake/Real]
<Explanation>: [Provide a detailed explanation based on your analysis of the article’s content, highlighting any indicators that led you to your decision.]\\
\bottomrule
\end{tabular}
\caption{Prompt for experiment of the Sequential Image-First.}
\label{tab:prompt_first_image}
\end{table*}

\begin{table}[h!]
\centering
\small
\begin{tabular}{@{}llc@{}}
\toprule
\textbf{Model} & \textbf{Tool} & \textbf{Count} \\ \midrule
\multirow{2}{*}{GPT Framework}& OpenAI ChatGPT\footnotemark[1] & 211  \\
 & Microsoft Copilot\footnotemark[2]   & 8\\ \midrule
Gemini Framework  & Google Gemini\footnotemark[3]   & 25   \\ \midrule
Mistral Framework & Mistral AI\footnotemark[4]  & 1    \\ \midrule
Unspecified & -   & 11   \\ \bottomrule
\end{tabular}
\caption{Distribution of model frameworks and tools used for fake news generation. Note: participants could use more than one tool. }
\label{tab:tools}
\end{table}
\footnotetext[1]{https://openai.com/chatgpt/}
\footnotetext[2]{https://copilot.microsoft.com/}
\footnotetext[3]{https://gemini.google.com/}
\footnotetext[4]{https://mistral.ai/}

\subsection{Robustness Check for Post-hoc Pairwise tests}
\label{sec:robustness}
Despite observing slight non-normality inherent in the count data, we proceeded with the ANOVA and parametric pairwise test due to its effectiveness in visualizing interaction effects and its intuitive interpretability. To validate the robustness of our findings, the analysis was replicated using non-parametric tests for all pairwise comparisons, which reaffirmed the observed patterns, as shown in Table~\ref{tab:pairwise_non_parametric} in the Appendix.
\subsection{Detailed Process of Fake News Topic Extraction}
\label{sec:clustering}
We employed topic modeling techniques using BERTopic for clustering and KeyBERT for generating topic representations. Building on topics identified through topic modeling, we further conducted a comprehensive thematic analysis of all 252 stories. This process included verifying the correct topic assignments for clustered stories and assigning unclustered stories to existing or new topics as needed. The model initially identified 7 distinct topics, while around 40 stories were left unclustered. Next, two independent coders with relevant expertise performed a manual thematic analysis and a second-stage review was conducted to refine the topic assignments and uncover any additional themes. Through this manual iterative process, an additional $8^{th}$ topic was created to categorize certain unclustered stories. However, even after manual review, 21 stories could not be grouped into any of these topics (e.g., because of low quality, etc.) and were therefore excluded from the analysis. 

\subsection{LLM-generated Indicators}

In the LLM fake news detection tasks, we asked the models to generate explanations that provided key reasons for its classification decisions. Among the three tested temperature settings, we selected the explanation output from the configuration that yielded the highest accuracy. We then used phrasemachine \cite{handler2016bag} to identify key phrases appearing more than three times in the itemized explanations.

The scatter plots presented in Figure~\ref{fig:scatter_gpt}, \ref{fig:scatter_gemini}, and \ref{fig:scatter_llama} in the Appendix present the frequency of these key phrases for GPT-4o, Gemini, and Llama-3.1, respectively, using Scattertext\footnote{https://github.com/JasonKessler/scattertext.git}. Phrases appearing further to the right on the x-axis are more frequently associated with news incorrectly identified as real by the LLM, and those higher on the y-axis are more frequently associated with news correctly identified as fake by the LLM. For GPT-4o, we observe that fake news containing contextual information, such as examples and details, is more difficult for the model to detect correctly (contextual information related terms are highlighted in red in all three figures). The keywords such as "contexts" and "details" frequently appears in the explanations for those mistakenly identified as real news, indicating that the presence of detailed context can obscure detection by the model. In the case of Gemini, while the term "context" also appears in misclassifications, additional narrative styles such as "consistent narrative" seem to play a role in deceiving the model (narrative style-related terms are highlighted in blue). These patterns indicate that Gemini falls short when it comes to recognizing fake news that mimics the structure and tone of real, neutral reporting. Open-source models like Llama-3.1, with limited capabilities and constrained parameter settings, exhibit sub-optimal performance in detecting fake stories, with misclassification often influenced by factors such as contextual enhancement and authority referencing (authority-related terms are highlighted in green).
These findings are consistent with prior observations that GPT-4o may be lacking when it pertains to detecting fake news when contextual enhancement is used, while Gemini performs poorly when narrative imitation is employed, where fake news mimics the structure and tone of real narratives.

\subsection{Comparison between AI-generated fake news with human instruction and with minimal instruction}

Through a detailed examination of the fake news creation process, we found that while Generative AI plays a significant role in generating fake news, it is not the only tool involved. Humans also make considerable efforts in iteratively refining the stories to enhance their believability. This leads us to question whether the fake news created with substantial human intervention is more challenging for LLMs to detect compared to purely AI-generated fake news with minimal human input.
To answer this, we evaluate the detection performance of LLMs on \emph{LLMFake} \citep{chen2023can}, a politics-related fake news dataset generated using ChatGPT-3.5. 
Since we are only interested in AI-generated fake news with minimal human intervention, we use only part of the datasets on fake news generated using \emph{Hallucinated News Generation} and \emph{Partially Arbitrary Generation} approaches. 
In total, $MI_{LLMFake}$ contains 300 politics-related fake news.
Furthermore, because this dataset is politics-related, we also extracted the politics-related news from all 252 fake news stories($HI_{Politics}$) for a more focused comparison beyond the overall dataset.
Table~\ref{tab:text_only} compares the detection accuracy of LLMs on fake news datasets exclusively generated by AI ($MI_{LLMFake}$), and datasets generated involving human-AI collaborations (i.e., $HI_{All}$ and $MI_{LLMFake}$).
We found that when humans are involved in the creation of fake news and employ various strategies to optimize the output, it becomes significantly more challenging for LLMs to detect these stories (on average 26\% lower in accuracy).

\begin{table}[h!]
\centering
\small
\begin{tabular}{cccc}
\toprule
 & \textbf{$HI_{All}$}&\textbf{$HI_{Politics}$} & \textbf{$MI_{LLMFake}$}  \\  
\midrule
GPT-4o &60.32\% & 64.71\%&\cellcolor{green!30}77.00\% \\
Gemini &68.65\% & 58.82\%&\cellcolor{green!30}80.00\% \\
Llama&39.29\%& 47.06\%&\cellcolor{green!30}72.67\%\\

\bottomrule
\end{tabular}%

\caption{Comparison of detection accuracy amongst all AI-generated fake news with human instruction, politics-related AI-generated fake news with human instruction, and PolitiFact fake news dataset with minimal instruction for all three models. $HI_{All}$: Detection accuracy for all AI-generated fake news with human instruction; $HI_{Politics}$: Detection accuracy for politics-related AI-generated fake news with human instruction; $MI_{LLMFake}$: Detection accuracy for LLMFake fake news dataset with minimal instruction; Green indicates better performance for each model. }
\label{tab:text_only}
\end{table}

\begin{figure*}[h!]
    \centering
    \includegraphics[scale=0.48]{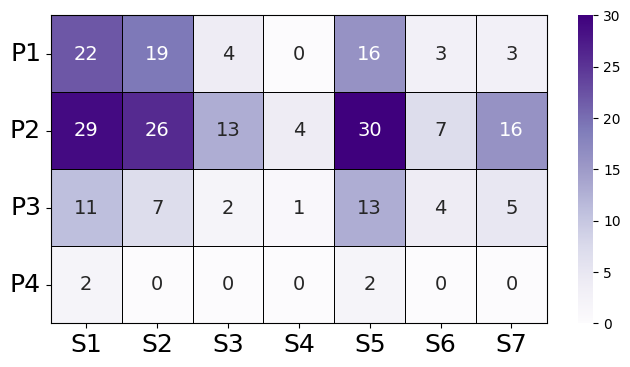}
    \caption{Co-usage heatmap of primary prompting strategies (P1-P4) and secondary output optimization strategies (S1-S7).  Note: Multiple primary and secondary strategies can be applied simultaneously. For simplicity, we only map the relationships between primary and secondary strategies. }
    \label{fig:HM}
\end{figure*}

\begin{figure*}[h!]
    \centering
    \includegraphics[scale=0.9]{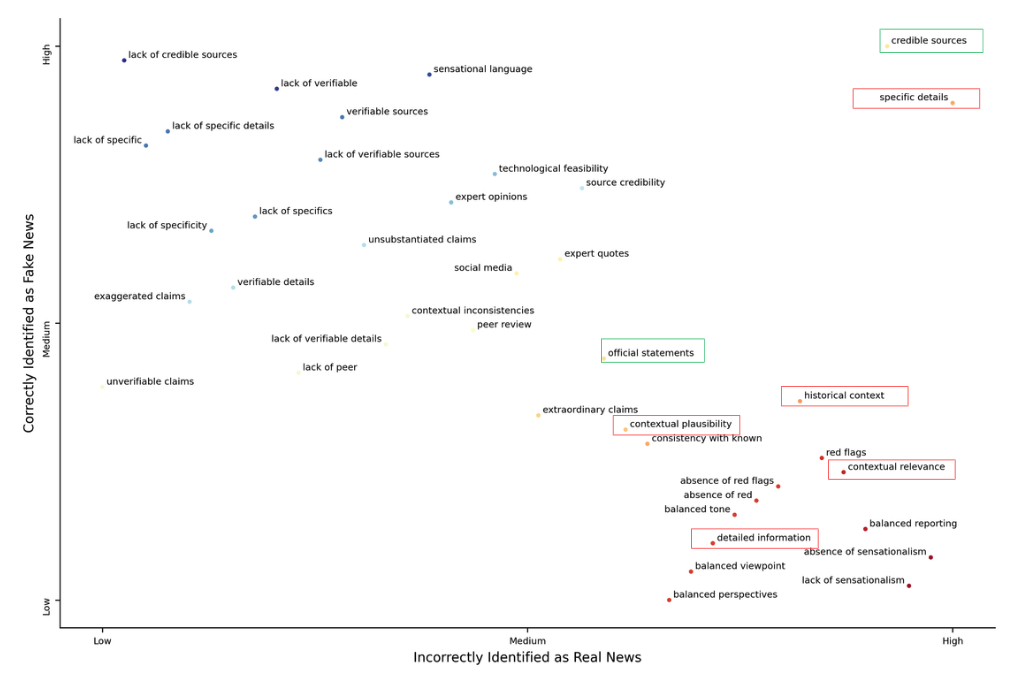}
    \caption{Scatter plot displaying the frequency of occurrence of terms in the indicators generated by GPT-4o ($Temp_{best}$=0.5), with the x-axis representing terms more frequently associated with news incorrectly identified as real by GPT-4o, and the y-axis representing terms more frequently associated with news correctly identified as fake by GPT-4o.}
    \label{fig:scatter_gpt}
\end{figure*}

\begin{figure*}[h!]
    \centering
    \includegraphics[scale=0.87]{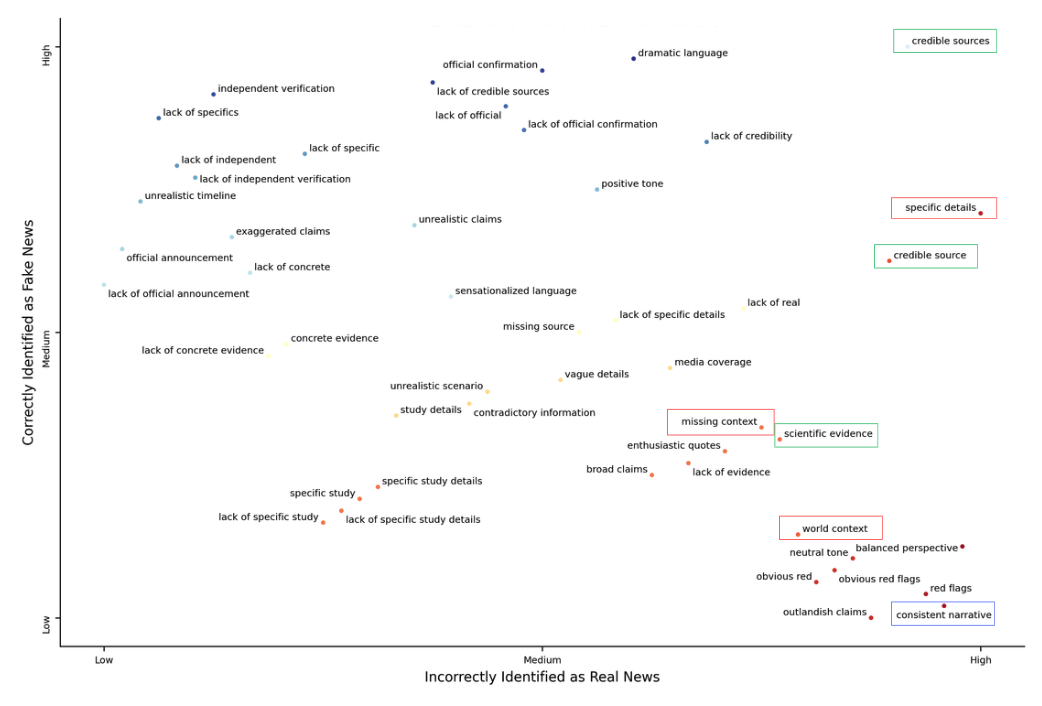}
    \caption{Scatter plot displaying the frequency of occurrence of terms in the indicators generated by Gemini ($Temp_{best}$=0.3), with the x-axis representing terms more frequently associated with news incorrectly identified as real by Gemini, and the y-axis representing terms more frequently associated with news correctly identified as fake by Gemini.}
    \label{fig:scatter_gemini}
\end{figure*}

\begin{figure*}[h!]
    \centering
    \includegraphics[scale=0.86]{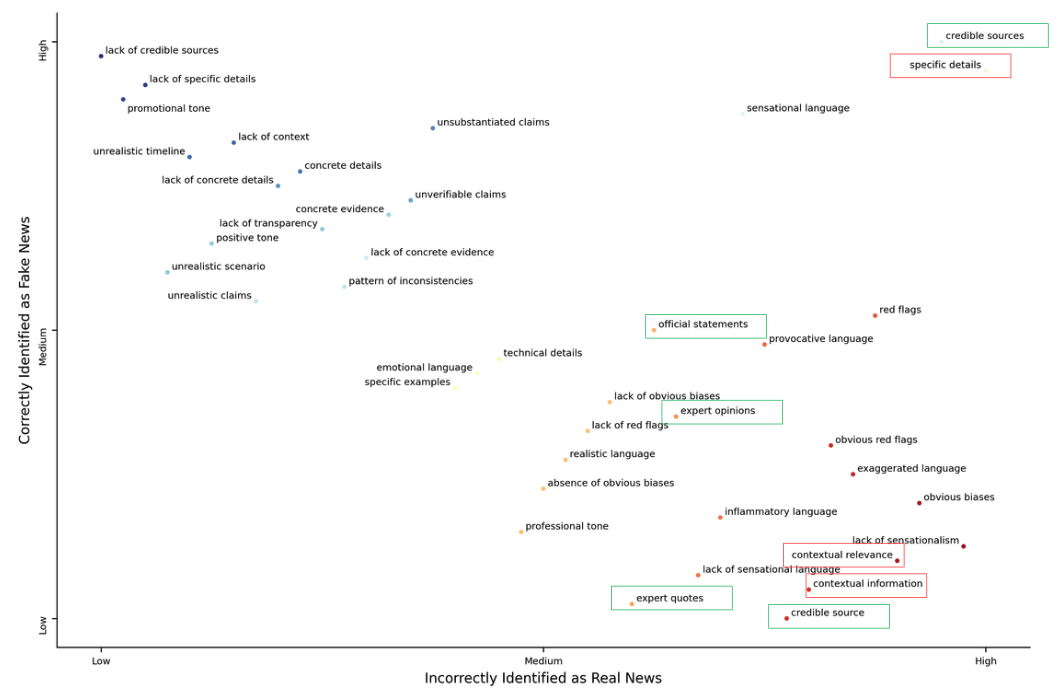}
    \caption{Scatter plot displaying the frequency of occurrence of terms in the indicators generated by Llama-3.1 ($Temp_{best}$=0.3), with the x-axis representing terms more frequently associated with news incorrectly identified as real by Llama-3.1, and the y-axis representing terms more frequently associated with news correctly identified as fake by Llama-3.1.}
    \label{fig:scatter_llama}
\end{figure*}

\end{document}